\newif\ifpreprint
\newif\ifauthorversion

\preprinttrue

\ifpreprint
\documentclass[sigconf,screen,nonacm]{acmart}
\else
\ifauthorversion
\documentclass[sigconf,screen,authorversion]{acmart}
\else
\documentclass[sigconf]{acmart}
\fi
\fi

\usepackage{lipsum}
\usepackage{soul}
\usepackage{color}

\DeclareRobustCommand{\hlpink}[1]{{\definecolor{foo}{HTML}{FFDBE6}\sethlcolor{foo}\hl{#1}}}
\DeclareRobustCommand{\hlblue}[1]{{\definecolor{foo}{HTML}{B3DBFF}\sethlcolor{foo}\hl{#1}}}
\DeclareRobustCommand{\hlorange}[1]{{\definecolor{foo}{HTML}{FFDFC1}\sethlcolor{foo}\hl{#1}}}
\DeclareRobustCommand{\hlgreen}[1]{{\definecolor{foo}{HTML}{CCEECC}\sethlcolor{foo}\hl{#1}}}

\begin{document}

\title[Beyond Summarization: Designing AI Support for Real-World Expository Writing Tasks]{Beyond Summarization: \\ Designing AI Support for Real-World Expository Writing Tasks}

\author{Zejiang Shen$^\dagger$, Tal August$^\square$, Pao Siangliulue$^\square$, Kyle Lo$^\square$, Jonathan Bragg$^\square$ \\ Jeff Hammerbacher$^\square$, Doug Downey$^{\square, \diamond}$, Joseph Chee Chang$^\square$, David Sontag$^\dagger$}
\affiliation{\institution{\vspace{1.5mm}$^\dagger$Massachusetts Institute of Technology, $^\square$Allen Institute for AI, $^\diamond$Northwestern University\vspace{1.5mm}}\country{}}
\email{{zjshen, dsontag}@mit.edu, {tala, paos, kylel, jbragg, jeffhammerbacher, dougd, josephc}@allenai.org}

\renewcommand{\shortauthors}{Shen, et al.}

\newcommand{\TODO}[1]{\textbf{\textcolor[rgb]{.8, .1, .1}{TODO: #1}}}
\newcommand{\ZS}[1]{\textbf{\textcolor[rgb]{.1, .8, .6}{SS: #1}}} 

\newcommand{\todo}[1]{\textcolor[rgb]{.8, .1, .1}{#1}}
\newcommand{\zs}[1]{\textcolor[rgb]{.1, .8, .6}{#1}} 
\newcommand{\ds}[1]{\textbf{\textcolor[rgb]{.8, .1, .8}{DS: #1}}}
\newcommand{\dd}[1]{\textbf{\textcolor[rgb]{.3, .3, .6}{DD: #1}}}
\newcommand{\jh}[1]{\textbf{\textcolor[rgb]{.7, .4, .6}{JH: #1}}}
\newcommand{\jb}[1]{\textbf{\textcolor[rgb]{.1, .1, .8}{JB: #1}}}
\newcommand{\ta}[1]{\textbf{\textcolor[rgb]{.1, .5, .3}{TA: #1}}}
\newcommand{\ps}[1]{\textbf{\textcolor[rgb]{.9, .6, .2}{PS: #1}}}
\newcommand{\joseph}[1]{\textbf{\textcolor[rgb]{.2, .6, .9}{JC: #1}}}

\begin{abstract}
  Large language models have introduced exciting new opportunities and challenges in designing and developing new AI-assisted writing support tools. 
  Recent work has shown that leveraging this new technology can transform writing in many scenarios such as ideation during creative writing, editing support, and summarization. 
  However, AI-supported \emph{expository writing}---including real-world tasks like scholars writing literature reviews or doctors writing progress notes---is relatively understudied.
  
  In this position paper, we argue that developing AI supports for expository writing has unique and exciting research challenges and can lead to high real-world impacts. 
  We characterize expository writing as evidence-based and knowledge-generating: it contains summaries of external documents as well as new information or knowledge. 
  It can be seen as the product of authors' sensemaking process over a set of source documents, and the interplay between reading, reflection, and writing opens up new opportunities for designing AI support. 
  We sketch three components for AI support design and discuss considerations for future research. 

\end{abstract}


\maketitle

\begin{figure*}[ht]
  \centering
    \includegraphics[width=0.75\linewidth]{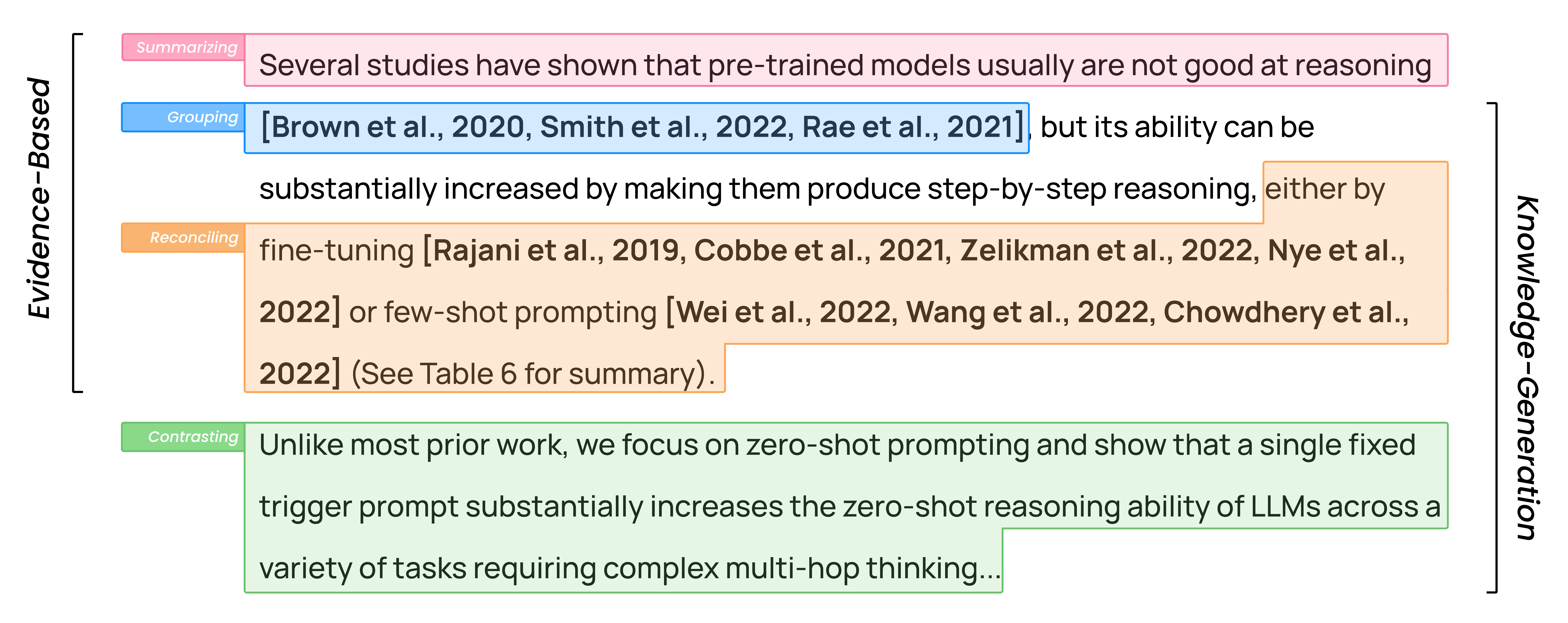}
    \caption{Dissecting an expository writing piece. {\normalfont In this example, the writing is a paragraph of a paper's related work section~\cite{kojima2022large}. While it contains \hlpink{summaries} of relevant previous work, it also elects to \hlblue{group} several papers together as they describe similar concepts, or \hlorange{reconcile} different findings under a shared framework. Finally, it \hlgreen{contrasts} the current paper with the previous work. The writing is \textbf{based on evidence} from existing work, but also \textbf{presents new knowledge} about the relationship between existing work as well as the novelty of the current paper, which are the final product of the authors' \textbf{sensemaking} process during the writing.}} 
    \label{fig:example-writing}
    \vspace{5mm}
\end{figure*}

\section{Introduction}

The advent of large language models (LLMs)~\cite{brown2020gpt3,Smith2022UsingDA, Rae2021ScalingLM} has brought about a dramatic change in the design space of AI-assisted writing.\footnote{In the remainder of this work, unless otherwise specified, AI-assisted writing refers to the use of LLMs to support writing.} 
The language understanding capabilities and high-quality text generation of LLMs promise to semi-automate cognitively-demanding writing tasks, i.e., help produce outlines or even generate long and grammatically correct paragraphs based on a short natural language input prompt.\footnote{\url{https://platform.openai.com/examples}}
As a result, there is growing interest from both the research and commercial communities in 
exploring new designs for intelligent writing support systems, including supporting creative story writing~\cite{lee2022coauthor, singh2022hide, chung2022talebrush}, blog posts or email composition,\footnote{For example, commercial tools like copy.ai, \url{https://copy.ai} and Lex, \url{https://lex.page/}.} personal knowledge management,\footnote{Including apps like mem, \url{https://get.mem.ai} and Notion-AI, \url{https://www.notion.so/}.} and so on. 

While prior work has explored many exciting applications for LLMs, we argue that \textbf{expository writing} is a task that is understudied in existing work on AI augmented writing.
We define expository writing pieces as articles that summarize facts and produce new knowledge or information. 
For example, this could be researchers collecting and reading multiple papers to write a survey paper~\cite{palani2023relatedly}, %
or doctors reviewing clinical notes to devise a treatment plan~\cite{kind2008documentation}. In these cases, authors not only summarize the source documents but also add information or bring new insights that do not exist in the source, e.g., organizing the relevant papers or synthesizing the patient's symptoms and test results to create a differential diagnosis and possible treatment options.

Compared to other types of writing that do not involve external documents, e.g., creative story or argumentative writing~\cite{lee2022coauthor, yang2022re3}, expository writing requires authors to comprehend the source, generate new insights, and faithfully reference and represent extracted information in the final article. 
The interaction between reading and writing brings new challenges for developing AI support, and the design is largely unexplored for incorporating the latest generations of language models. 
While summarization is often a necessary component, expository writing is also distinguished from document summarization tasks in terms of its goal to bring about new information that does not exist in the source.

Expository writing can be seen as a \textbf{sensemaking} process~\cite{russell1993cost}, and different types of sub-tasks are involved: typically, authors start with iteratively exploring and reading multiple relevant documents to \emph{identify and extract key evidence}, then they \emph{organize the evidence into useful schema} and further \emph{synthesize into coherent writing} to communicate new knowledge or information~\cite{Pirolli2007TheSP}.
Therefore, the role of AI may vary during the process of writing, and we argue it is important to design different types and levels of AI support to maximally help the authors while minimally influencing or shaping their opinions. 
In one approach, in the early stages of writing, the writers would initiate and drive the work and AI should only provide limited supporting functions.
As ideas manifest and authors have a better sense of the writing content, AI could assume more responsibilities with authors ``supervising'' the model's work. 
For example, when starting writing a survey paper, the authors come up with the query to find relevant papers first, and AI helps execute the search and discover related documents; after authors read the retrieved papers and come up with ideas for the writing, AI can help generate the writing text based on authors' ideas, and the writers only need to proofread model output.
AI discovery helps the user learn better, and editing support makes the writing more efficient and polished; most importantly, writers are in full control of the thinking and the ideas included in the produced articles.

Expository writing occurs frequently in many real-world tasks, and we argue that 
{\em real-world expository writing}
is a high value open problem for AI augmented writing support, with many challenges and opportunities in the space. 
Many realistic expository writing tasks require domain experts, e.g., describing key events in lawsuits~\cite{shen2022multi}, explaining scientific concepts or ideas~\cite{King2022DontSW}, and briefing on patients' conditions and treatments~\cite{adams21_whats_summar}.  
Successful augmented writing systems for these tasks stand to both reduce the expensive expert hours required to perform the writing, and to improve the quality of the output (e.g. \citet{bell2020frequency} reports that 1 in 5 patients find a mistake in their clinical notes written by doctors or nurses). 
Often, supporting such real-world tasks also allows us to draw upon existing rich repositories of example data and established evaluation protocols, which can further the development of future AI augmented writing systems.
In the following sections, we formally define the expository task and sketch components of the design that assist
the writing process rather than just the final article.

\section{Characterizing Expository Writing}

In contrast to other writing tasks, expository writing has two unique characteristics: it is \emph{evidence-driven} and \emph{knowledge-generating} as illustrated in Figure~\ref{fig:example-writing}.
Formally, given a set of \emph{source documents} and a collection of \textit{writing objectives}, expository writing aims to compose a \emph{piece} that synthesizes the information from the source and produces \emph{new information} in accordance with the objectives.
There are different ways for authors to synthesize the source, including but not limited to the \emph{selection}, \emph{grouping}, \emph{contrasting}, or \emph{reconciliation} of multiple pieces of similar or different (even conflicting) statements, and the authors' synthesis brings information not present in the source documents. 
For example, a doctor might \emph{select and group} a set of relevant observations from clinical notes, and \emph{reconcile} with a possible condition to achieve the goal of \emph{devising} a treatment plan. 
The writing objectives guide both the reading and synthesis process, and they can also be updated given new insights generated during the course of writing. 
Unlike a summarization task, here, it is not necessary to involve all source documents in the final writing: in fact, the choice to not include a document also constitutes new information in the writing product (i.e., by indicating the document's relative importance to the goal).

\section{Designing AI Supports for Expository Writing} 

To optimally involve LLMs to help expository writing, we argue that there are three components: 
1) supporting the reading and evidence-seeking for correct and efficient understanding of the source, 2) assisting information synthesis and the production of new knowledge and ideas, and 3) facilitating text composition to communicate relevant evidence and insights. 
Expository writing pieces aim at bringing about new knowledge or perspectives, but it takes significant effort for authors to comprehend the source and convert the thinking into writing. 
By reducing the cognitive load and the interaction costs for information extraction and sensemaking during the reading, it can help authors focus on the reasoning and idea generation~\cite{Kang2022Threddy,palani2023relatedly}. 
On the other hand, while LLMs have demonstrated strong performance in document understanding and text generation, they currently suffer from hallucination~\cite{ji2022survey} and are limited in reasoning~\cite{brown2020gpt3, Smith2022UsingDA} and long-form generation capabilities.
As such, it is sub-optimal to use LLMs to generate the whole writing piece altogether and post-edit~\cite{cheng2022mapping}.
A modular design that provides varying AI support at different stages of writing can be most helpful, and we detail the components as follows.  

\paragraph{Augmenting Reading and Collecting Evidence} 
Reading relevant documents to gather evidence is a crucial early step for expository writing. 
While there have been significant research efforts on document discovery and comprehension (e.g., for scientific documents~\cite{Cohan2020SPECTERDR,Rachatasumrit2022CiteReadIL,Chang2022CiteSee,Kang2023ComLitteeLD,Head2020AugmentingSP}), AI tools for supporting \emph{reading for writing} is relatively sparse~\cite{Han2022PassagesIW,Kang2022Threddy,Rachatasumrit2021ForSenseAO}.
Experts typically expend significant manual effort reading through many source documents to identify key relevant information to help them synthesize new knowledge.
This is referred as ``establishing the working memory''~\cite{hayes2000new,baddeley1992working}, a key step during the cognitive process of writing.
One existing approach to facilitate reading is using language models to automatically generate summaries for long documents~\cite{brown2020gpt3, zhang23_bench_large_languag_model_news_summar}, but they can also be prone to hallucination~\cite{ji2022survey}.
More importantly, \citet{ziegler2019fine} shows that recent instruction-tuned models are merely ``smart copiers'' when performing summarization, and overly trusting the model outputs risks biasing authors' views and may lead to authors collecting inaccurate evidence.

As such, authors' reading and understanding the source still plays a central role, and designing for \emph{reading for expository writing} should focus on the \emph{augmentation} of reading and extraction of key information from source documents \cite{Kang2022Threddy}, but at the same time providing support for verifying the extracted evidence.
For example, the system can automatically generate a list of the key facts relevant to an author's goal from a set of source documents, and guide the authors to examine and check the facts' correctness by linking each one back to relevant parts of the original source documents. 

\paragraph{Supporting Information Synthesis} 

As an important middle step between reading and writing, the writer typically takes a long time to inspect and reason over the evidence collected and produce new ideas to be written, e.g., finding the similarities and differences between papers and reconciling them under a shared framework as illustrated in Figure~\ref{fig:example-writing}.
Synthesizing a large of collection of information gathered across source documents can be challenging, and different interfaces and techniques have been developed to support this step~\cite{Han2022PassagesIW,Rachatasumrit2021ForSenseAO,cattan2021scico}. 
Language models can be involved to improve the efficiency and accuracy of these approaches by, e.g., automatic grouping based on the textual representation~\cite{reimers2019sentence} or enabling semantic search.
Beyond better organizing evidence to reflect an author's mental model, a system could also leverage LLMs for inspiration during synthesis \cite{Gero2021SparksIF}. 
One example is proposing connections between evidences and suggesting relevant/different topics to discover, though it is important to also provide mechanisms so that authors can verify and ground model-generated ideas with relevant evidence. 
The involvement of AI models is not intended to be a replacement for authors' thinking and contemplation but enhancement for producing new ideas, and discovering otherwise unseen connections or findings. 

\paragraph{Facilitating Text Composition} 
The translation of ideas into language has been considered to be equally (if not more) cognitively demanding and challenging as the other components: as \citet{flower1981cognitive} puts, it needs ``juggle all the special demands of written English'' and ``the process can interfere with ... planning''. 
The latest language models have demonstrated strong capabilities in generating fluent and high-quality text, and some studies suggest the generations are even on par with freelance writers for certain summarization tasks~\cite{zhang23_bench_large_languag_model_news_summar}.
It is promising to incorporate language models to support text composition, but several design choices need to be considered to reduce the risk of generation errors and biasing authors' opinions~\cite{mirowski2022co}.
For example, it may be important for language models in this case to be evidence-aware and explicit---the generation should only be invoked when the authors call it, and relevant evidence and a short prompt about the writing intent needs to be provided.
An instance of model generation could be relatively short, e.g., one or two sentences, with the authors calling it in multiple turns. 
Compared to generating long completions at one time, e.g., a whole paragraph, we speculate that this could decrease the possibility of erroneous and biased generation, and decrease the cost of verification by the authors. 

\section{Conclusion}

Expository writing is genre of evidence-driven and knowledge-generating writing that takes place in many real-world settings. 
In this paper, we argue that the unique characteristics in expository writing open up new opportunities for designing AI support.
We sketch the components of the design and highlight considerations and challenges for future implementation.   

\section*{Acknowledgments}

We thank the anonymous reviewers for their insightful comments and suggestions. We appreciate the helpful discussion and advice from Oren Etzioni, Arvind Satyanarayan, Dennis Wei, Prasanna Sattigeri, Subhro Das, Barbara Lam, Lauren Yu, and Ruochen Zhang.

\bibliographystyle{ACM-Reference-Format}
\bibliography{main}

\end{document}